\documentclass[10pt, a4paper]{article}

\usepackage[final]{lrec2026} 

\usepackage{threeparttable}
\usepackage{booktabs}
\usepackage{arydshln}
\usepackage{array}
\usepackage{graphicx}
\usepackage{caption,subcaption}
\usepackage{xcolor}

\usepackage{tikz}

\usetikzlibrary{positioning,arrows.meta,decorations.pathreplacing,shadows,fit,calc,backgrounds}

\usepackage{fontawesome5}

\definecolor{richgreen}{RGB}{240, 50, 50}
\definecolor{slateblue}{RGB}{60,105,200}

\newcommand{\chapicon}{\hspace{0.19em}\raisebox{0.0em}{\textcolor{slateblue}{\faBook}}\hspace{0.42em}}
\newcommand{\hiericon}{\raisebox{0.0em}{\textcolor{richgreen}{\faSitemap}}\hspace{0.2em}}

\newcommand{\chaplabel}{%
  \makebox[0.8em][c]{\raisebox{-1.2ex}{\chapicon}}}

\newcommand{\hierlabel}{%
  \makebox[0.8em][c]{\raisebox{-1.2ex}{\hiericon}}}

\makeatletter
\def\adl@drawiv#1#2#3{%
        \hskip.5\tabcolsep
        \xleaders#3{#2.5\@tempdimb #1{1}#2.5\@tempdimb}%
                #2\z@ plus1fil minus1fil\relax
        \hskip.5\tabcolsep}
\newcommand{\cdashlinelr}[1]{%
  \noalign{\vskip\aboverulesep
           \global\let\@dashdrawstore\adl@draw
           \global\let\adl@draw\adl@drawiv}
  \cdashline{#1}
  \noalign{\global\let\adl@draw\@dashdrawstore
           \vskip\belowrulesep}}
\makeatother

\usepackage{amsmath, amssymb}
\usepackage{cleveref}

\usepackage[most]{tcolorbox}

\tcbset {
  base/.style={
    arc=2mm,
    bottomtitle=0.5mm,
    boxrule=0.1mm,
    colbacktitle=black!10!white, 
    coltitle=black, 
    fonttitle=\bfseries, 
    left=2.5mm,
    right=3.5mm,
    title={#1},
    toptitle=0.75mm, 
  }
}

\tcbuselibrary{listings}

\definecolor{systemcolor}{HTML}{6BA4D9}
\definecolor{usercolor}{HTML}{7CC5AD}
\definecolor{assistantcolor}{HTML}{E57F84}
\definecolor{revisioncolor}{HTML}{D9A76B}

\tcbset{
    custombox/.style={
        enhanced,
        colback=white,
        colframe=gray!20,
        boxrule=0.5pt,
        fonttitle=\bfseries,
        coltitle=black,
        attach boxed title to top left={yshift=-2mm, xshift=2mm},
        boxed title style={size=small, colback=white},
        separator sign none,
    }
}

\definecolor{variablebg}{RGB}{247,247,247}
\definecolor{variableborder}{RGB}{227,227,227}
\definecolor{variabletext}{RGB}{204,0,0}

\newcommand{\formatvariable}[1]{%
    \tcbox[
        on line,
        size=fbox,
        colback=variablebg,
        colframe=variableborder,
        arc=1mm,
        boxrule=0pt,
        left=0.5pt,
        right=0.5pt,
        top=0pt,
        bottom=0pt
    ]{\textcolor{variabletext}{\ttfamily\small#1}}%
}

\usepackage{siunitx}
\title{\textit{Paragraph Segmentation Revisited}:\\Towards a Standard Task for Structuring Speech}

\name{Fabian Retkowski$^1$, Alexander Waibel$^{1,2}$} 

\address{\textsuperscript{1}Karlsruhe Institute of Technology, \textsuperscript{2}Carnegie Mellon University \\
         \texttt{retkowski@kit.edu}, \texttt{waibel@cmu.edu}}

\abstract{
Automatic speech transcripts are often delivered as unstructured word streams that impede readability and repurposing. We recast paragraph segmentation as the missing structuring step and fill three gaps at the intersection of speech processing and text segmentation. First, we establish \textsc{TEDPara} (human-annotated TED talks) and \textsc{YTSegPara} (YouTube videos with synthetic labels) as the first benchmarks for the paragraph segmentation task. The benchmarks focus on the underexplored speech domain, where paragraph segmentation has traditionally not been part of post-processing, while also contributing to the wider text segmentation field, which still lacks robust and naturalistic benchmarks. Second, we propose a constrained-decoding formulation that lets large language models insert paragraph breaks while preserving the original transcript, enabling faithful, sentence-aligned evaluation. Third, we show that a compact model (MiniSeg) attains state-of-the-art accuracy and, when extended hierarchically, jointly predicts chapters and paragraphs with minimal computational cost. Together, our resources and methods establish paragraph segmentation as a standardized, practical task in speech processing. 
 \\ \newline  }

\begin{document}

\maketitleabstract

\section{Introduction}

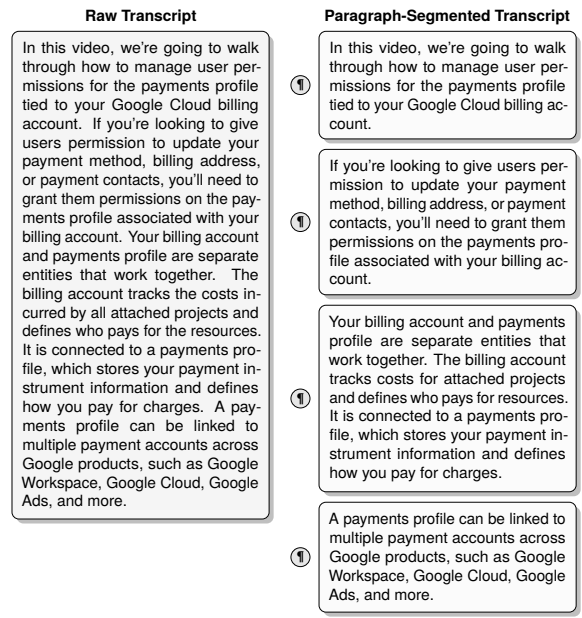
\begin{figure}[t]
\centering
\resizebox{\columnwidth}{!}{
\begin{tikzpicture}[font=\footnotesize, >=Stealth, node distance=4mm and 6mm]
\tikzset{
  card/.style={draw, rounded corners, fill=gray!8, inner sep=6pt, text width=4.8cm, align=justify, drop shadow},
  cardbg/.style={draw, rounded corners, fill=gray!8, inner sep=6pt, text width=4.8cm, align=justify},
  para/.style={draw, rounded corners, fill=gray!3, inner sep=6pt, text width=4.8cm, align=justify, drop shadow},
  badge/.style={circle, draw, fill=gray!20, inner sep=0pt, minimum size=11pt, font=\bfseries\scriptsize}
}

\node[card] (raw) {%
In this video, we're going to walk through how to manage user permissions for the payments profile tied to your Google Cloud billing account. If you're looking to give users permission to update your payment method, billing address, or payment contacts, you'll need to grant them permissions on the payments profile associated with your billing account. Your billing account and payments profile are separate entities that work together. The billing account tracks the costs incurred by all attached projects and defines who pays for the resources. It is connected to a payments profile, which stores your payment instrument information and defines how you pay for charges. A payments profile can be linked to multiple payment accounts across Google products, such as Google Workspace, Google Cloud, Google Ads, and more.
};

\coordinate (rightTop) at ($ (raw.north east) + (10mm,0) $);

\node[para, anchor=north west] (p1) at (rightTop) {%
In this video, we're going to walk through how to manage user permissions for the payments profile tied to your Google Cloud billing account.};
\node[para, below=2.0mm of p1.south west, anchor=north west] (p2) {%
If you're looking to give users permission to update your payment method, billing address, or payment contacts, you'll need to grant them permissions on the payments profile associated with your billing account.};
\node[para, below=2.0mm of p2.south west, anchor=north west] (p3) {%
Your billing account and payments profile are separate entities that work together. The billing account tracks costs for attached projects and defines who pays for resources. It is connected to a payments profile, which stores your payment instrument information and defines how you pay for charges.};
\node[para, below=2.0mm of p3.south west, anchor=north west] (p4) {%
A payments profile can be linked to multiple payment accounts across Google products, such as Google Workspace, Google Cloud, Google Ads, and more.};

\begin{scope}[on background layer]
\node[cardbg, fill opacity=0, draw opacity=0, fit={(p1)(p2)(p3)(p4)}, inner sep=6pt] (segbox) {};
\end{scope}

\node[anchor=south west, font=\bfseries\footnotesize, text depth=0pt]
  (rawhdr) at ($ (raw.north west) + (39pt,2pt) $) {Raw Transcript};

\node[anchor=south west, font=\bfseries\footnotesize] at ($ (segbox.north west) + (6pt,-6pt) $) {Paragraph-Segmented Transcript};

\node[badge, left=1.7mm of p1.west] (b1) {¶};
\node[badge, left=1.7mm of p2.west] (b2) {¶};
\node[badge, left=1.7mm of p3.west] (b3) {¶};
\node[badge, left=1.7mm of p4.west] (b4) {¶};

\end{tikzpicture}
}
\caption{Paragraph segmentation turns an undifferentiated transcript into visually coherent paragraphs, aiding readability and navigation.}
\label{fig:paragraphs}
\end{figure}

Readability is a major concern for automatic speech recognition (ASR) transcripts, which are traditionally output as unstructured sequences of words, frequently lacking punctuation, casing, and higher-level organization such as sentence or paragraph boundaries \cite{jones_measuring_2003,shugrina_formatting_2010,tundik_user-centric_2018}. While much research has focused on restoring sentence-level structure, paragraph segmentation is a less explored area. Paragraphs help users navigate and understand long-form speech such as lectures or meetings, where ideas unfold over extended spans \cite{lai_automatic_2016}. They also improve the usability and visual clarity of transcripts, avoiding the appearance of a dense, unreadable wall of text (\Cref{fig:paragraphs}). Human evaluations show a preference for paragraph-segmented transcripts with breaks enhancing comprehension and perceived coherence \cite{pappu_automatic_2015}.

Despite its importance, paragraph segmentation remains underexplored in speech processing, in part due to the absence of standardized benchmarks and the scarcity of labeled data for spoken content. Unlike sentence boundary detection or punctuation restoration, paragraph segmentation lacks large-scale, curated datasets, making systematic evaluation and model development difficult.

To address this gap, we introduce \textsc{TEDPara} and \textsc{YTSegPara}, the first benchmarks for paragraph segmentation in speech, covering both TED talks and YouTube videos. Beyond that, these datasets fill a key void in the broader text segmentation area, where high-quality spoken-domain benchmarks remain scarce. We provide strong baselines, including a compact model (MiniSeg) and LLM–based methods, and propose an efficient constrained decoding approach that inserts paragraph breaks while preserving transcript fidelity, essential for faithful evaluation. Using \textsc{YTSegPara}, we further extend MiniSeg to hierarchical modeling, enabling the joint prediction of chapter and paragraph boundaries, demonstrating that both levels can be learned efficiently within a shared framework. Evaluation combines automatic metrics with human judgments, providing a comprehensive perspective on segmentation quality. Together, these contributions establish a foundation for treating paragraph segmentation as a standardized, measurable task in speech processing and highlight the practical value of our datasets for future research.

\section{Related Work}

\paragraph{Paragraph Segmentation in Written Text.} Paragraph segmentation has largely been explored in the context of written text, where paragraph boundaries are typically already available. As a result, it is often treated as a secondary task, commonly used in self-supervised pretraining for sentence segmentation \cite{wicks_unified_2021,minixhofer_wheres_2023,frohmann_segment_2024}, rather than as a primary research objective. Only a few recent studies have directly addressed paragraph segmentation as their main focus, typically within specific domains such as news articles and literary texts \cite{iikura_improving_2021,zhuo_auxiliary_2023,yoo_improving_2024}, but these efforts remain isolated without a standardized task definition or benchmark.

\paragraph{Segmentation of Speech Transcripts.} Research on segmentation in speech data has been limited. Early work on video transcripts explored automatic paragraph segmentation strategies \cite{lai_automatic_2016,salimbajevs_system_2017,lai_integrating_2020}, but these efforts did not produce reusable benchmarks, thereby limiting reproducibility and broader applicability. Recent work has focused on higher-level segmentation, targeting chapter segmentation in long-form spoken content with joint title generation \cite{ghazimatin_podtile_2024,retkowski_text_2024}. While related, these approaches address a coarser segmentation granularity and involve broader objectives.

\paragraph{Text Segmentation Benchmarks.}

Current research in text segmentation is constrained by the scarcity of high-quality datasets. As noted by \citet{glavas_training_2021}, the field suffers from an ``absence of large annotated datasets,'' a limitation reflected in earlier work that often relied on small datasets or benchmarks constructed by concatenating unrelated snippets (e.g., \citealt{lukasik_text_2020}). A recent survey \cite{ghinassi_recent_2024} identifies this lack of suitable resources as the central challenge for progress. To date, only a few large-scale benchmarks exist, most prominently \textsc{Wiki-727K} \cite{koshorek_text_2018} and more recently \textsc{YTSeg} \cite{retkowski_text_2024}, which focus on topic- or chapter-level segmentation. In contrast, we introduce paragraph segmentation in spoken transcripts as a distinct task at a finer granularity. By establishing dedicated benchmarks for this setting, we broaden the empirical landscape of segmentation research and enable more diverse and robust model evaluation across tasks and datasets under the shared framework of text segmentation.

\paragraph{Constrained Decoding with LLMs.} Constrained decoding and structured output generation enable LLMs to produce outputs that follow specific formats or rules. Recent work has explored grammar-based and input-dependent constraints \cite{geng_grammar-constrained_2023,geng_flexible_2023}. However, these works do not consider paragraph boundaries as a structured prediction problem.

  
\section{Task Definition}

\textit{Paragraph segmentation} is a special case of \textit{text segmentation} whose goal is to divide a text into paragraph units. While there is no single agreed-upon definition of what constitutes a \textit{paragraph}, it is often described as a semantically or functionally coherent segment \cite{bolshakov_text_2001}. At the same time, boundaries may also be introduced for stylistic reasons, considering discourse structure and rhetorical roles \cite{sporleder_automatic_2004}, transitional and connective phrases \cite{zadrozny_semantics_1991,lai_automatic_2016}, or length and readability \cite{yoo_improving_2024}. Paragraph segmentation operates on a finer granularity compared to \textit{topic segmentation}, which predicts higher-level topic shifts that also typically imply paragraph breaks \cite{filippova_using_2006}.

\section{Dataset Construction}

\subsection{TEDPara}

\textsc{TEDPara} is derived from publicly available TED Talk transcripts, which include high-quality, human-annotated paragraph structure aligned with spoken presentations. We restrict our dataset to English transcripts only and collect all TED Talks listed on the official TED website as of May 13, 2024, spanning content published since February 2006. This results in an initial pool of 6,379 talks.

\paragraph{Preprocessing.} We apply the following filtering steps. First, we remove all talks that lack a transcript, affecting 724 talks (12.8\%). Next, we exclude talks that contain only a single paragraph, as they do not provide any paragraph boundary information; this step removes an additional 462 talks (7.2\%). The final \textsc{TEDPara} dataset contains 5,193 talks with multi-paragraph transcripts, which we randomly partitioned into training, validation, and testing splits; see \Cref{table:tedpara_split} for details.


  \begin{table*}[h!]
  \small
  \centering
  \begin{tabular}{lrrrrrrr}
      \toprule
\textbf{Split}
& \textbf{\# Talks}
& \textbf{\# Sent.}
& \textbf{\# Para.}
& \textbf{Sent./Talk}
& \textbf{Para./Talk}
& \textbf{Sent./Para.}
& \textbf{Breaks (\%)} \\
      \midrule
      Train & 4{,}154 {\small (80\%)} & 467{,}255 & 106{,}719 & 112.5 & 25.7 & $4.4 \pm 4.0$ & 22.0 \\
      Val   &    519  {\small (10\%)} &  60{,}257 &  13{,}697 & 116.1 & 26.4 & $4.4 \pm 4.7$ & 21.9 \\
      Test  &    520  {\small (10\%)} &  60{,}212 &  13{,}534 & 115.8 & 26.0 & $4.5 \pm 4.1$ & 21.6 \\      \midrule
      Total & 5{,}193 & 587{,}724 & 133{,}950 & 113.2 &  25.8 & $4.4 \pm 4.1$ &  21.9 \\
      \bottomrule
  \end{tabular}
  \caption{Data splits and dataset statistics for \textsc{TEDPara}}
  \label{table:tedpara_split}
  \end{table*}

\begin{table}[t]
\small
\centering
\begin{tabular}{lrrr}
\toprule
\textbf{Dataset} & \textbf{Doc Len.} & \textbf{\# Seg./Doc} & \textbf{Seg. Len.} \\
\midrule
\textsc{TEDPara}  & 113.2 & 25.8 & 4.4 \\
\textsc{YTSeg}    & 196.2 & 9.12 & 21.5 \\
\textsc{Wiki-727K}& 57.6  & 3.48 & 13.6 \\
\bottomrule
\end{tabular}
\caption{Segmentation granularity comparison across large-scale text segmentation datasets.}
\label{table:seg_granularity}
\end{table}

\paragraph{Dataset Statistics.} Table~\ref{table:seg_granularity} shows that \textsc{TEDPara} targets a finer segmentation level than large-scale benchmarks such as \textsc{YTSeg} and \textsc{Wiki-727K}. While these datasets contain fewer segments per document (9.12 and 3.48) with longer segments (21.5 and 13.6 sentences/segment), \textsc{TEDPara} has 25.8 paragraphs per talk with 4.4 sentences/paragraph on average. Together with its intermediate document length (113.2 sentences/talk), \textsc{TEDPara} complements existing benchmarks by expanding coverage in both granularity and document length.

\paragraph{Release.} Due to licensing restrictions on TED content, we do not redistribute the data directly. Instead, we provide both the partitioned talk IDs as well as scripts for downloading and preprocessing the talks into a standardized format\footnote{\url{https://github.com/retkowski/tedseg}}, ensuring both reproducibility and legal compliance.

\subsection{YTSegPara}

To extend our task to more diverse speech content, we augment the existing \textsc{YTSeg} dataset \citelanguageresource{ytseg}, which provides chapter annotations for structurally and topically diverse YouTube videos. The original dataset is limited to English-language videos with English transcripts, and the transcripts are derived from closed captions, which lack paragraph structure. We augment the dataset with paragraph-level annotations to produce a new dataset: \textsc{YTSegPara}.

\begin{table}[h]
  \small
  \centering
  \begin{tabular}{lrr}
  \toprule
  \textbf{Dataset} & \textbf{Para./Doc} & \textbf{Sent./Para.} \\
  \midrule
  \textsc{TEDPara}   & 25.8 & 4.4 \\
  \textsc{YTSegPara} & 44.6 & 4.2 \\
  \bottomrule
  \end{tabular}
  \caption{Paragraph granularity comparison between \textsc{TEDPara} and \textsc{YTSegPara}.}
  \label{table:para_granularity}
  \end{table}

\paragraph{Augmentation.} Since manual paragraph annotation is infeasible at scale, we derive paragraph boundaries using the LLM-based constrained decoding method described in \Cref{sec:llm_paragraphic}, using the LLaMA 3.1 70B model \cite{grattafiori_llama_2024}.

\paragraph{Dataset Statistics.} Table~\ref{table:para_granularity} compares the paragraph granularity of the two datasets. Both exhibit a similar paragraph density of  $\approx 4$ sentences per paragraph. However, \textsc{YTSegPara} contains nearly twice as many paragraphs per document (44.6 vs.\ 25.8), reflecting its longer documents.

\paragraph{Utility and Scope.} Paragraph segmentation is a narrow problem that can be reduced to a sequence of binary decisions, making it well-suited for knowledge distillation into smaller, more efficient models (e.g., a Transformer encoder classifier) where inductive biases can effectively be imposed. Thus, the generated annotations serve a dual purpose: they provide training data for compact models and establish a benchmark for hierarchical segmentation, jointly predicting both chapters and paragraphs, in long-form spoken content.

\paragraph{Release.} \textsc{YTSegPara} inherits the original dataset’s CC BY-NC-SA 4.0 license and is released with scripts and metadata\footnote{\url{https://github.com/retkowski/ytsegpara}}.

\section{Paragraph Insertion with LLMs}
\label{sec:llm_paragraphic}

\begin{figure}[!ht]
\begin{center}
\includegraphics[width=\columnwidth]{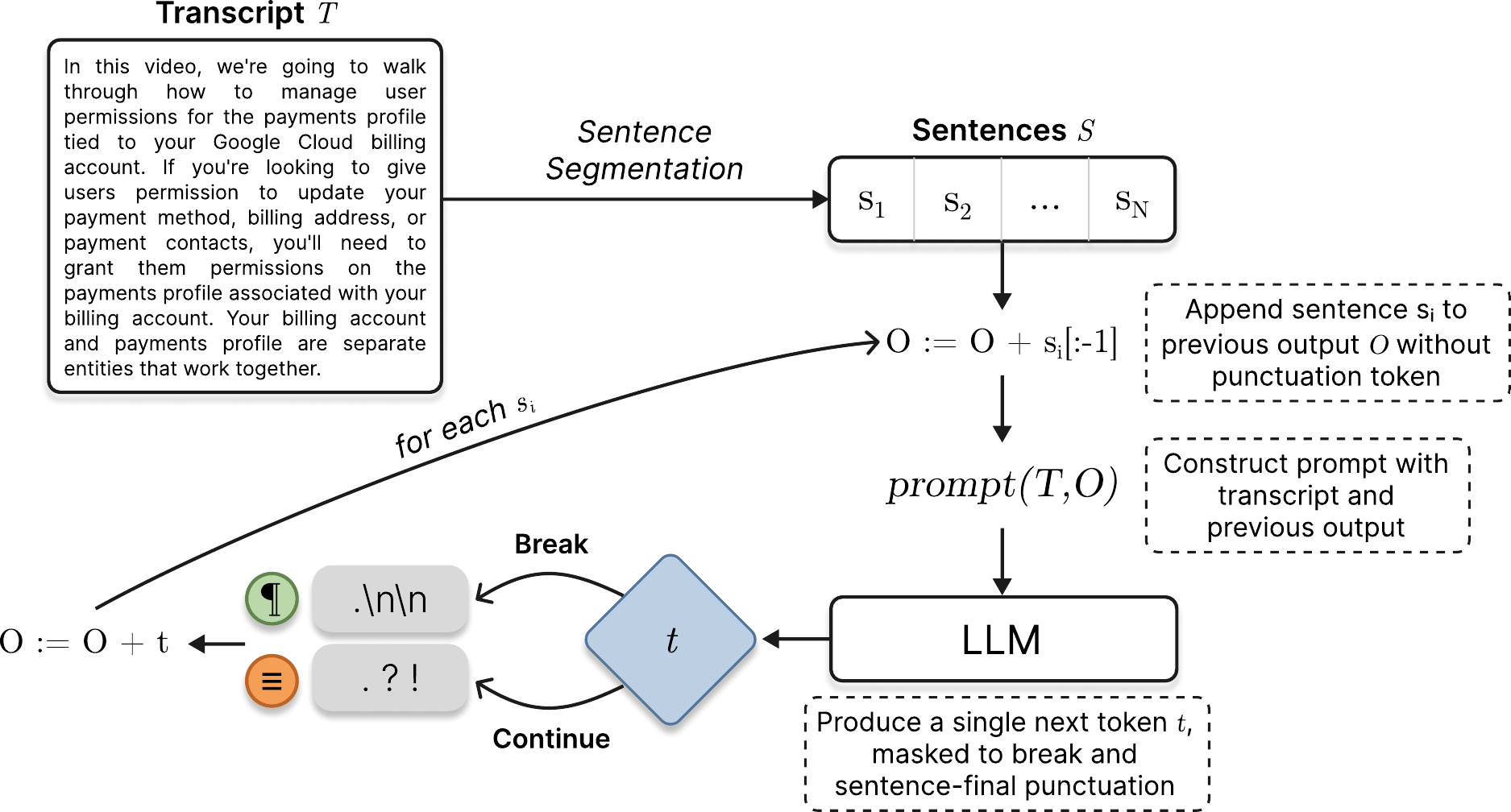}
\caption{Conceptual illustration of the sentence-wise constrained decoding algorithm for paragraph insertion with LLMs. At each sentence boundary, the model performs a single forward pass to decide between two constrained actions: \textit{\colorbox[HTML]{f49f55}{continue}} (emit standard punctuation) or \textit{\colorbox[HTML]{bad8aa}{break}} (emit punctuation followed by newlines).}
\label{fig:constrained_decoding}
\end{center}
\end{figure}

We cast paragraph segmentation as a \emph{constrained completion} task (\Cref{fig:constrained_decoding}):
at each sentence boundary the LLM may emit \emph{only}
(i) the punctuation token that ends the current sentence (``continue''), or
(ii) punctuation tokens followed by the delimiter
``\texttt{\textbackslash n\textbackslash n}'' (``break'').
Because the model cannot hallucinate arbitrary text, the original
transcript is preserved verbatim and the paragraph structure emerges from
a sequence of binary choices.

\paragraph{Procedure.} Formally, let $T$ be the transcript and $S=(s_1,\dots,s_m)$ its sentence segmentation, obtained with NLTK's sentence tokenizer \cite{bird_natural_2009}. 
Before boundary $i$ we hold the partially formatted output $O$, whose last token is the sentence-final punctuation $p\in P$, where $P$ is the set of sentence-ending punctuation tokens (e.g., \texttt{.}, \texttt{!}, \texttt{?}).
We strip $p$ (as a form of \textit{token healing}; \citealt{lundberg_ribeiro_2023_token_healing}) and build a prompt from the transcript $T$ and the output prefix $O$.
We tokenize this prompt as $x=\tau(\mathrm{prompt}(T,O))$, where $\tau$ is the model tokenizer, and query the $\mathsf{LLM}$ for the next-token distribution $\mathbf p=\mathsf{LLM}(x)$, i.e., $\mathbf p[t]=p_{\mathsf{LLM}}(t\mid x)$.
Let $N$ be the set of allowed \emph{break} tokens (punctuation tokens followed by the delimiter \texttt{\textbackslash n\textbackslash n}). We mask all other tokens and compare:

\begin{align}
p_{\text{punct}} &= p_{\mathsf{LLM}}(p \mid x), \\
p_{\text{break}} &= \max_{t \in N} p_{\mathsf{LLM}}(t \mid x).
\end{align}

If $p_{\text{break}}>p_{\text{punct}}$ we append the most likely break token $t^\star=\arg\max_{t\in N}p_{\mathsf{LLM}}(t\mid x)$ and set $\pi_{i+1}=1$ (paragraph break); otherwise we restore $p$ and keep $\pi_{i+1}=0$.
Finally, we copy the next sentence $s_{i+1}$ verbatim to $O$, adjusting whitespace to avoid consecutive spaces.

\paragraph{Efficiency.} The loop performs exactly one forward pass of $\mathsf{LLM}$ per sentence boundary, for an $O(m)$ runtime; far cheaper than the $O(|\tau(T)|)$ generations a full re-write would need.

\paragraph{Ensuring Comparable Evaluation.} Unconstrained decoding often modifies content or structure, producing outputs that differ from the reference transcript. Such deviations undermine evaluation reliability because automatic metrics in text segmentation including F1, $P_k$ and Boundary Similarity depend on consistent sentence boundaries. If transcripts change, results become incomparable across systems.

\paragraph{Applicability.} Importantly, constrained decoding is orthogonal to the choice of zero-shot versus fine-tuned inference. A task-adapted LLM benefits equally from the constrained formulation: it retains the efficiency of a single forward pass per boundary and guarantees verbatim transcript preservation.

\paragraph{Section-Wise Processing.} When a dataset already provides coarser units such as chapters (as in \textsc{YTSeg}), we process each block independently. This ensures that predicted paragraph boundaries are coherent with the higher-level structure, while keeping inputs within the context window and enabling work on hierarchical segmentation.

\paragraph{Generalization Across Tokenizers.} We observe two families of tokenization behavior with respect to paragraph delimiters. In the first, exemplified by LLaMA~3 \cite{grattafiori_llama_2024} and Qwen~2.5 \cite{qwen25} the tokenizer merges sentence-ending punctuation and the delimiter into a single compound token (e.g., \texttt{.\textbackslash n\textbackslash n}). In the second, exemplified by Gemma~3 \cite{gemma3}, the delimiter \texttt{\textbackslash n\textbackslash n} is encoded as a separate token. The compound case is handled by the token-healing procedure described above. For the separate-token case, we can simplify it to a \emph{next-word} variant: rather than stripping the final punctuation, we let the model score the first token of the following sentence and compare it against tokens that begin with \texttt{\textbackslash n\textbackslash n}. The decision rule is identical; the model chooses between continuing and breaking and only the token sets differ.

\paragraph{Release.} For reproducibility and documentation, we publish the inference scripts with their decoding algorithm and prompts under a CC-BY 4.0 license\footnote{\url{https://github.com/retkowski/paragraph-llm}}.

\section{Experiments}
\subsection{Experimental Setup}

Our experiments aim to establish strong baselines for our newly introduced benchmarks, \textsc{TEDPara} and \textsc{YTSegPara}, and to provide insights into the relationship between chapter segmentation and paragraph segmentation, the effectiveness of hierarchical segmentation, and the quality of LLM labeling used for pseudo-annotated data.

\paragraph{Models.} In our experiments, we utilize the LLaMA 3 series, specifically LLaMA 3.1 8B, LLaMA 3.1 70B, and LLaMA 3.3 70B \cite{grattafiori_llama_2024} as examples for LLM-based paragraph segmentation, both in zero-shot and fine-tuned variants (LoRA). At the same time, we report results with MiniSeg \cite{retkowski_text_2024}, as a model shown to have a strong performance in chapter segmentation of speech transcripts.

\paragraph{Hierarchical Segmentation.} For experiments on \textsc{YTSegPara}, we extend MiniSeg to \textit{hierarchical segmentation} by framing the segmentation on different levels (chapter-level and paragraph-level) as a \textit{multi-class classification} problem.

\paragraph{Metrics.} We report three metrics: F1 score, Boundary Similarity (BS; \citealt{fournier_evaluating_2013}), and $P_k$ \cite{beeferman_statistical_1999}. F1 score is a classic classification metric that is increasingly used for text segmentation due to its interpretability. BS is a proposed and promising metric that captures graded boundary similarity. $P_k$ is a well-established standard for evaluating segmentation tasks.

\paragraph{Baselines.} Baselines include random and rule-based segmentations, multiple variants of MiniSeg trained with different pretraining and fine-tuning regimes, e.g., pretrained with \textsc{Wiki-727K} \citelanguageresource{wiki727k} as well as the LLaMA 3 series as an example for LLM-based segmentation.

\paragraph{Random Baseline.} For TEDPara, we sample paragraph boundaries uniformly at random within each transcript, while assuming oracle knowledge of the reference number of paragraph breaks per document; this ensures the baseline matches the true break count but not their locations. For YTSegPara, we similarly sample chapter boundaries uniformly at random given the oracle number of chapters, then place paragraph breaks within each chapter span using the global paragraph-break rate.

\paragraph{Rule-Based Baseline.} We include a simple deterministic baseline that inserts paragraph breaks at a fixed interval. Specifically, we place a boundary after every $n$ sentences, where $n$ is set to the global average sentences per paragraph in the corresponding split (rounded). For TEDPara, this gives $n=5$ for both validation ($4.57$) and test ($4.63$).

\paragraph{Human Evaluation.} For human evaluation, we used two complementary methods. Pairwise comparisons were aggregated into ELO ratings, an increasingly common way to quantify relative preferences in a single measure \cite{boubdir_elo_2023,chiang_chatbot_2024}, while Likert-scale judgments (1–5) provided absolute quality assessments. This combination captures both relative and absolute perspectives on segmentation quality.

\paragraph{Human Evaluation Methodology.} We conducted a two-part study on the TEDPara test split with 8 participants (4 per subtask), each completing 30 judgements. In Subtask 1, annotators performed randomized, blind A/B comparisons of paragraph segmentations with three response options (A, B, tie); presentation order and sampling were randomized to mitigate position bias and ensure balanced model coverage. In Subtask 2, annotators rated single segmentations on a 5-point Likert scale (1 = Poor, 5 = Excellent), using a similar balanced sampling procedure. Across both subtasks, we compared random and rule-based baselines, LLM-generated segmentations (with and without PBR), and human-annotated references.

\paragraph{Token-Weighted LoRA Fine-Tuning.} Paragraph breaks are sparse at the sentence level and vanishingly so at the token level, where newline tokens are vastly outnumbered by content tokens. While adjusted losses have been explored for encoder-based paragraph segmentation \cite{iikura_improving_2021,retkowski_text_2024}, these operate on per-sentence classification. We investigate a simple analogue for autoregressive LLM fine-tuning: upweighting the newline token in the language modeling loss during training.

\subsection{Results and Discussion}

\begin{table}[h]
\centering
\small
\begin{threeparttable}
\resizebox{\columnwidth}{!}{%
\begin{tabular}{lcccc}
\toprule
\textbf{Model} & \textbf{Exact} (↑) & \textbf{+Whitespace} (↑) & \textbf{+Punct./Case} (↑) & \textbf{Len. ±5\%} (↑) \\
\midrule
\multicolumn{5}{c}{\textbf{Na\"ive Decoding}} \\
\cdashlinelr{1-5}
LLaMA 3.1 8B       & 0.59 & 0.80 & 0.83 & 0.99 \\
LLaMA 3.1 70B      & 0.64 & 0.87 & 0.88 & 1.00 \\
LLaMA 3.3 70B      & 0.63 & 0.89 & 0.91 & 1.00 \\
\midrule
\multicolumn{5}{c}{\textbf{Constrained Decoding}} \\
\cdashlinelr{1-5}
LLaMA 3.1 8B       & 1.00 & 1.00 & 1.00 & 1.00 \\
LLaMA 3.1 70B      & 1.00 & 1.00 & 1.00 & 1.00 \\
LLaMA 3.3 70B      & 1.00 & 1.00 & 1.00 & 1.00 \\
\bottomrule
\end{tabular}%
}
\end{threeparttable}
\caption{Divergence of generated output from the original \textsc{TEDPara} transcript when using unconstrained decoding. Reported are proportions of outputs matching the input exactly (ignoring only paragraph breaks), then progressively relaxing constraints to ignore whitespace, punctuation/casing, and smaller length variations.}
\label{tab:llm_hallucinations}
\end{table}

\paragraph{Hallucination Risks with LLMs.} As demonstrated in \Cref{tab:llm_hallucinations}, unconstrained decoding frequently introduces hallucinations, even under relaxed matching criteria. Although ignoring paragraph breaks, whitespace, and punctuation/casing improves match rates, divergence persists in a meaningful number of cases (ranging from 0.09 to 0.17, depending on the model). In the most lenient setting, which only requires outputs to be within 5\% of the original length, LLaMA 3.1 8B still falls short of perfect fidelity ($0.99$), indicating occasional severe hallucinations such as dropped text parts. These results strongly motivate the constrained decoding approach introduced in this paper, which not only improves efficiency but also ensures faithful preservation of the input transcript.

\begin{table}[h]
\centering
\small
\begin{threeparttable}
\resizebox{\columnwidth}{!}{%
\begin{tabular}{lcccccc}
\toprule
\textbf{Model} & \multicolumn{3}{c}{\textbf{Val}} & \multicolumn{3}{c}{\textbf{Test}} \\
\cmidrule(r){2-4} \cmidrule(l){5-7}
& \textbf{F1} (↑) & \textbf{BS} (↑) & \textbf{$P_k$} (↓) 
& \textbf{F1} (↑) & \textbf{BS} (↑) & \textbf{$P_k$} (↓) \\
\midrule
\multicolumn{7}{c}{\textbf{Zero-Shot}} \\
\cdashlinelr{1-7}
 Random Baseline & 16.8 & 15.9 & 49.7 & 17.1 & 15.8 & 49.6 \\
 Rule-Based Baseline & 21.6 & 24.2 & 49.9 & 22.4 & 24.0 & 49.8 \\

 LLaMA 3.1 8B & 36.6 & 30.5 & 38.7 & 33.9 & 28.3 & 40.3 \\
 LLaMA 3.1 70B & 42.8 & 35.6 & 37.5 & 40.7 & 33.9 & 38.5 \\
 LLaMA 3.3 70B & 37.5 & 30.1 & 37.7 & 37.1 & 29.3 & 38.2 \\
MiniSeg \scriptsize(Wiki) & 24.6 & 14.9 & 37.6 & 25.0 & 15.3 & 38.0 \\
MiniSeg \scriptsize(Wiki)\small\tnote{a} & 30.5 & 20.3 & 36.3 & 29.7 & 19.6 & 28.3 \\ 
 MiniSeg \scriptsize(Wiki $\rightarrow$ YT) & 33.2 & 21.4 & 36.1 & 32.7 & 21.5 & 36.7 \\
 MiniSeg \scriptsize(Wiki $\rightarrow$ YT)\small\tnote{a} & 45.7 & 32.6 & 32.4 & 43.4 & 30.9 & 33.2 \\
 \midrule
 \multicolumn{7}{c}{\textbf{Zero-Shot + Paralinguistic Break Rule}} \\
  \cdashlinelr{1-7}
 LLaMA 3.1 8B + PBR\tnote{b} & 51.7 & 44.3 & 32.9 & 49.9 & 42.5 & 34.2 \\
 LLaMA 3.1 70B + PBR\tnote{b} & 55.5 & 47.6 & 32.4 & 54.0 & 46.7 & 33.1 \\
 LLaMA 3.3 70B + PBR\tnote{b} & 52.2 & 43.7 & 32.3 & 52.5 & 43.5 & 32.5 \\
\midrule
\multicolumn{7}{c}{\textbf{Fine-Tuned on \textsc{TEDPara}}} \\
\cdashlinelr{1-7}
 LLaMA 3.1 8B \scriptsize(LoRA; TED) & 69.7 & 58.6 & 21.9 &  68.4 & 57.2 & 22.7 \\
 MiniSeg \scriptsize(TED) & 67.3 & 56.1 & 24.3 & 67.3 & 56.3 & 24.0 \\
 MiniSeg \scriptsize(YT $\rightarrow$ TED) & 69.8 & 60.2 & 22.4 & 70.6 & 61.2 & 21.6 \\
 MiniSeg \scriptsize(Wiki $\rightarrow$ TED) & 70.6 & 60.7 & 21.8 & 71.2 & 61.5 & 21.0 \\
 MiniSeg \scriptsize(Wiki $\rightarrow$ YT $\rightarrow$ TED) & \textbf{72.1} & \textbf{62.2} & \textbf{21.0} & \textbf{72.7} & \textbf{63.2} & \textbf{20.1} \\
\bottomrule
\end{tabular}%
}
\begin{tablenotes}
\scriptsize
\begin{minipage}{0.935\columnwidth}
\item[a] Threshold tuned across partitions ($\tau_{\text{val}}{=}0.264$, $\tau_{\text{test}}{=}0.300$ for Wiki; $\tau_{\text{val}}{=}0.257$, $\tau_{\text{test}}{=}0.293$ for Wiki $\rightarrow$ YT).
\item[b] Paralinguistic Break Rule (PBR): Additional, rule-based post-processing to insert paragraph breaks around standalone paralinguistic cues.
\end{minipage}
\end{tablenotes}
\end{threeparttable}
\caption{Performance comparison of baselines for paragraph segmentation on \mbox{\textsc{TEDPara}}, using different approaches and (pre-)training strategies.}
\label{tab:model_results}
\end{table}

\paragraph{From Chapters to Paragraphs.} The results in \Cref{tab:model_results} highlight a strong connection between chapter and paragraph segmentation. The zero-shot performance of models trained on chapter-level data improves notably when the segmentation threshold is lowered, leading to more frequent segment predictions and outputs that better align with paragraph structure. Additionally, pretraining on data with higher-level segments such as \textsc{Wiki-727K} and \textsc{YTSeg} leads to strong results when fine-tuned on \textsc{TEDPara}, showing effective transfer of structural cues between domains and segmentation levels. Overall, the findings confirm that pretraining on related segmentation tasks significantly benefits paragraph segmentation.

\paragraph{Paralinguistic Parentheticals.} We conducted a qualitative investigation to better understand the performance gap between the fine-tuned models and the LLM-based approaches. One consistent pattern we identified is that the \textsc{TEDPara} reference annotations reliably place paralinguistic parentheticals such as "\textit{(Laughter)}" or "\textit{(Applause)}" in their own paragraphs. In contrast, LLMs treat these as inline elements. This discrepancy introduces a systematic mismatch that is not due to a fundamental limitation of the LLMs, but rather the absence of a simple formatting rule. Once this rule is applied by inserting paragraph boundaries before and after standalone paralinguistic cues, the performance gap narrows, as can be seen in \Cref{tab:model_results}.

\begin{table}[h]
\centering
\small
\begin{threeparttable}
\resizebox{\columnwidth}{!}{%
\begin{tabular}{lcccccc}
\toprule
\textbf{Model} & \multicolumn{3}{c}{\textbf{Paragraph Seg.}} & \multicolumn{3}{c}{\textbf{Chapter Seg.}} \\
\cmidrule(r){2-4} \cmidrule(l){5-7}
 & \textbf{F1} (↑) & \textbf{BS} (↑) & \textbf{$P_k$} (↓) & \textbf{F1} (↑) & \textbf{BS} (↑) & \textbf{$P_k$} (↓) \\
\midrule
Random Baseline & 26.3 & 26.6 & 51.3 & 7.6 & 8.7 & 47.9 \\
\chapicon\ MiniSeg \scriptsize(Wiki $\rightarrow$ TED)\tnote{a} & 35.7 & 32.2 & 43.6 & -- & -- & -- \\
\chapicon\ MiniSeg \scriptsize(Wiki) & -- & -- & -- & 12.5 & 8.8 & 40.5 \\

\chapicon\ MiniSeg \scriptsize(Wiki $\rightarrow$ YT) & -- & -- & -- & 46.1 & 38.2 & 27.0 \\

\hiericon\ MiniSeg \scriptsize(YT) & 47.9 & 42.2 & 33.4 & 43.6 & 32.5 & 28.8 \\

\hiericon\ MiniSeg \scriptsize(Wiki $\rightarrow$ YT) & 50.5 & 44.5 & 32.2 & 43.8 & 35.7 & 28.2 \\

\bottomrule
\end{tabular}%
}
\begin{tablenotes}
\scriptsize
\begin{minipage}{0.935\columnwidth}
\item[a] Predicted paragraph boundaries, scored within the oracle chapter spans.
\item[\chaplabel] Trained on either chapter or paragraph segmentation only.
\item[\hierlabel] Trained hierarchically on both paragraph and chapter segmentation.
\end{minipage}
\end{tablenotes}
\end{threeparttable}
\caption{Performance comparison of baselines for hierarchical segmentation (consisting of paragraph segmentation and chapter segmentation) on \mbox{\textsc{YTSegPara}}, using different variants of MiniSeg.}
\label{tab:para_chap_seg_results}
\end{table}

\paragraph{Efficient, Hierarchical Segmentation.} \Cref{tab:para_chap_seg_results} shows that adding the paragraph task to MiniSeg results in only minimal impact on chapter-level performance: chapter F1 decreases slightly from 46.1 to 43.8, and $P_k$ increases modestly from 27.0 to 28.2. This is notable given the increased inter-class confusion typically expected when expanding the label space. At the same time, the model produces a useful paragraph segmenter (paragraph F1 50.5). These results suggest that the existing parameter budget is sufficient to model both levels, yielding, to our knowledge, the first hierarchical segmentation model for speech and audiovisual transcripts.

\begin{table}[h]
\centering
\small
\resizebox{\columnwidth}{!}{%
\begin{tabular}{lcccccccc}
\toprule
\textbf{Token Weight} & \multicolumn{4}{c}{\textbf{Val}} & \multicolumn{4}{c}{\textbf{Test}} \\
\cmidrule(r){2-5} \cmidrule(l){6-9}
& \textbf{P} (↑) & \textbf{R} (↑) & \textbf{F1} (↑) & \textbf{\# Para.}
& \textbf{P} (↑) & \textbf{R} (↑) & \textbf{F1} (↑) & \textbf{\# Para.} \\
\midrule
1.0  & 74.9 & 65.1 & 69.7 & 21.7 & 74.1 & 63.5 & 68.4 & 20.6 \\
1.5            & 70.8 & 68.8 & 69.8 & 24.4 & 69.6 & 67.9 & 68.8 & 23.7 \\
2.0            & 67.2 & 72.5 & 69.7 & 27.0 & 66.2 & 73.5 & 69.7 & 27.3 \\
\cdashlinelr{1-9}
Reference      & --   & --   & --   & 26.4 & --   & --   & --   & 26.0 \\
\bottomrule
\end{tabular}%
}
\caption{Effect of newline token weighting during LoRA fine-tuning on \textsc{TEDPara}. Higher weights increase the number of predicted segments, raising recall at the cost of precision.}
\label{tab:token_weight}

\end{table}

\paragraph{Token-Weighted Fine-Tuning.} In our LoRA fine-tuning setup, the model tends to slightly undersegment relative to the reference (Table~\ref{tab:token_weight}). To explore whether this can be corrected, we experiment with upweighting the newline token in the language modeling loss during LoRA training. Increasing the weight encourages the model to predict paragraph breaks more frequently, effectively trading precision for recall. While this allows practitioners to calibrate the segmentation density to their needs, we find that adjusting the token weight does not improve the overall F1 score: gains in recall are offset by corresponding drops in precision.

\begin{table}[t]
\centering
\setlength{\tabcolsep}{3pt}
\begin{subtable}{0.48\columnwidth}
\centering
\tiny
\begin{tabular}{l r}
\toprule
\textbf{Model} & \textbf{ELO Score} \\
\midrule
LLaMA 3.1 70B + PBR & 1050.5 \\
LLaMA 3.1 70B       & 1034.9 \\
Reference            & 1015.9 \\
Rule-Based Baseline  & 1005.9 \\
Random Baseline      &  892.8 \\
\bottomrule
\end{tabular}
\subcaption{Pairwise preference evaluation using ELO.}
\label{tab:elo}
\end{subtable}%
\hfill
\begin{subtable}{0.48\columnwidth}
\centering
\tiny
\begin{tabular}{l r}
\toprule
\textbf{Model} & \textbf{Score} \\
\midrule
LLaMA 3.1 70B       & 3.64 $\pm$ 0.73 \\
LLaMA 3.1 70B + PBR & 3.55 $\pm$ 1.12 \\
Reference            & 3.50 $\pm$ 1.10 \\
Rule-Based Baseline  & 3.30 $\pm$ 0.98 \\
Random Baseline      & 2.88 $\pm$ 1.32 \\
\bottomrule
\end{tabular}
\subcaption{Likert-scale ratings of segmentation quality.}
\label{tab:likert}
\end{subtable}

\caption{Human evaluation results on \textsc{TEDPara} test data: (a) ELO scores from pairwise preference; (b) Likert-scale ratings (1 = Poor, 5 = Excellent).}
\label{tab:human_eval}
\end{table}

\paragraph{Human Validation of LLM Outputs.} As presented in \Cref{tab:human_eval}, pairwise comparisons on \textsc{TEDPara} yielded ELO ratings that place LLaMA 3.1 above the human reference and well above baselines, while Likert-scale judgments confirmed that its outputs are rated on par with references and consistently outperform rule-based or random baselines. These results are consistent with the automatic evaluation in \Cref{tab:model_results}, where LLM-based segmentation approaches the performance of supervised systems trained directly on reference segmentations. Together, these findings strengthen confidence that our approach produces paragraph structures comparable to human-authored segmentations and is suitable for use as a benchmark.

\paragraph{Rule-Based Baseline Performance.}
While rule-based segmentation performs poorly on automated metrics, its human evaluation scores are higher than expected given its simplicity. This likely reflects the visual structure it introduces: evenly spaced paragraph breaks reduce visual density and create a more readable layout. As \citet{stark_what_1988} argued, paragraphing often serves stylistic functions rather than marking clear linguistic or semantic boundaries. The resulting visual separation can thus produce a sense of coherence and intentionality even when breaks are not meaningfully placed.

\section{Conclusion}
This work lays the foundation for treating paragraph segmentation as a standardized and measurable task in speech processing by introducing two complementary benchmarks: \textsc{TEDPara} and \textsc{YTSegPara}. \textsc{TEDPara} provides a high-quality, human-annotated reference grounded in formal spoken presentations, while \textsc{YTSegPara} covers a broad spectrum of real-world spoken content with synthetic labels generated via constrained LLM decoding. Together, these datasets capture a range of speech domains and conditions, forming a robust foundation for training and evaluation, not only for paragraph segmentation of speech transcripts but also in the broader research area of text segmentation, which has notoriously lacked benchmarks.

Our proposed constrained decoding method enables LLMs to efficiently and faithfully insert paragraphs without altering the original transcript. While fine-tuned models achieve stronger automatic scores, human evaluations rate the LLM outputs on par with or above human references, showing their potential as pseudo-labels for benchmarking. In addition, experiments demonstrate that paragraph and chapter segmentation can be modeled jointly with minimal performance trade-offs, enabling efficient, hierarchical structuring of speech transcripts. These contributions not only address a longstanding gap in transcript formatting but also support downstream tasks, including summarization and information retrieval, offering practical value for applications in education, accessibility, and knowledge management.

\section{Limitations}

While our benchmarks and methods advance paragraph segmentation for spoken content, several limitations remain. First, \textsc{YTSegPara} relies on synthetic labels generated via constrained decoding, which may not fully align with human judgment. However, our human evaluation provides encouraging evidence that LLM-generated segmentations broadly align with human preferences. Second, we observed a systematic stylistic mismatch between model and reference conventions, particularly in the handling of paralinguistic cues, which affects automatic metrics despite comparable perceived quality. Third, our datasets focus on structured, relatively clean speech from TED talks and YouTube videos, leaving out more challenging domains such as conversational meetings, where noise and disfluencies are more prevalent. Finally, our current approach operates solely on textual transcripts. While this simplifies processing and broadens applicability, it misses potentially useful prosodic cues from the audio signal, such as pauses, pitch, and intonation, that could further improve segmentation quality.

\section{Acknowledgements}

This research is supported by the project "How is AI Changing Science? Research in the Era of Learning Algorithms" (HiAICS), funded by the Volkswagen Foundation. In addition, we acknowledge the computing time provided on the high-performance computer HoreKa by the National High-Performance Computing Center at KIT (NHR@KIT). This center is jointly supported by the Federal Ministry of Education and Research and the Ministry of Science, Research and the Arts of Baden-Württemberg, as part of the National High-Performance Computing (NHR) joint funding program. HoreKa is partly funded by the German Research Foundation (DFG).

\section{Bibliographical References}\label{sec:reference}

\bibliographystyle{lrec2026-natbib}
\bibliography{custom}

\begin{thebibliography}{2}
\expandafter\ifx\csname natexlab\endcsname\relax\def\natexlab#1{#1}\fi

\bibitem[{Koshorek et~al.(2018)Koshorek, Cohen, Mor, Rotman, and Berant}]{wiki727k}
Koshorek, Omri and Cohen, Adir and Mor, Noam and Rotman, Michael and Berant, Jonathan. 2018.
\newblock \href {https://github.com/koomri/text-segmentation} {\emph{Wiki-727K}}.
\newblock TAD - The Center for AI \& Data Science, Tel Aviv University. Distributed via GitHub.

\bibitem[{Retkowski and Waibel(2024)}]{ytseg}
Retkowski, Fabian and Waibel, Alexander. 2024.
\newblock \href {https://doi.org/10.57967/hf/1824} {\emph{YTSeg (2024-07-25)}}.
\newblock Interactive Systems Lab, Karlsruhe Institute of Technology. Distributed via Hugging Face.
\newblock PID \href{https://doi.org/10.57967/hf/1824}{https://doi.org/10.57967/hf/1824}.
\newblock License: CC-BY-NC-SA-4.0.

\end{thebibliography}


\begin{thebibliography}{36}
\expandafter\ifx\csname natexlab\endcsname\relax\def\natexlab#1{#1}\fi

\bibitem[{Beeferman et~al.(1999)Beeferman, Berger, and Lafferty}]{beeferman_statistical_1999}
Doug Beeferman, Adam Berger, and John Lafferty. 1999.
\newblock \href {https://doi.org/10.1023/A:1007506220214} {Statistical {Models} for {Text} {Segmentation}}.
\newblock \emph{Machine Learning}, 34(1):177--210.

\bibitem[{Bird et~al.(2009)Bird, Klein, and Loper}]{bird_natural_2009}
Steven Bird, Ewan Klein, and Edward Loper. 2009.
\newblock \emph{Natural language processing with {Python}}, 1st edition.
\newblock O'Reilly, Beijing ; Cambridge [Mass.].
\newblock OCLC: ocn301885973.

\bibitem[{Bolshakov and Gelbukh(2001)}]{bolshakov_text_2001}
Igor~A. Bolshakov and Alexander~F. Gelbukh. 2001.
\newblock \href {https://doi.org/10.1007/3-540-44805-5_20} {Text {Segmentation} into {Paragraphs} {Based} on {Local} {Text} {Cohesion}}.
\newblock In \emph{Text, {Speech} and {Dialogue}, 4th {International} {Conference}, {TSD} 2001, {Zelezna} {Ruda}, {Czech} {Republic}, {September} 11-13, 2001, {Proceedings}}, volume 2166 of \emph{Lecture {Notes} in {Computer} {Science}}, pages 158--166. Springer.

\bibitem[{Boubdir et~al.(2023)Boubdir, Kim, Ermis, Hooker, and Fadaee}]{boubdir_elo_2023}
Meriem Boubdir, Edward Kim, Beyza Ermis, Sara Hooker, and Marzieh Fadaee. 2023.
\newblock \href {https://aclanthology.org/2023.gem-1.28/} {Elo {Uncovered}: {Robustness} and {Best} {Practices} in {Language} {Model} {Evaluation}}.
\newblock In \emph{Proceedings of the {Third} {Workshop} on {Natural} {Language} {Generation}, {Evaluation}, and {Metrics} ({GEM})}, pages 339--352, Singapore. Association for Computational Linguistics.

\bibitem[{Chiang et~al.(2024)Chiang, Zheng, Sheng, Angelopoulos, Li, Li, Zhu, Zhang, Jordan, Gonzalez, and Stoica}]{chiang_chatbot_2024}
Wei-Lin Chiang, Lianmin Zheng, Ying Sheng, Anastasios~N. Angelopoulos, Tianle Li, Dacheng Li, Banghua Zhu, Hao Zhang, Michael~I. Jordan, Joseph~E. Gonzalez, and Ion Stoica. 2024.
\newblock Chatbot arena: an open platform for evaluating {LLMs} by human preference.
\newblock In \emph{Proceedings of the 41st {International} {Conference} on {Machine} {Learning}}, {ICML}'24. JMLR.org.
\newblock Place: Vienna, Austria.

\bibitem[{Filippova and Strube(2006)}]{filippova_using_2006}
Katja Filippova and Michael Strube. 2006.
\newblock \href {https://aclanthology.org/W06-1632/} {Using linguistically motivated features for paragraph boundary identification}.
\newblock In \emph{Proceedings of the 2006 {Conference} on {Empirical} {Methods} in {Natural} {Language} {Processing}}, pages 267--274, Sydney, Australia. Association for Computational Linguistics.

\bibitem[{Fournier(2013)}]{fournier_evaluating_2013}
Chris Fournier. 2013.
\newblock \href {https://aclanthology.org/P13-1167} {Evaluating {Text} {Segmentation} using {Boundary} {Edit} {Distance}}.
\newblock In \emph{Proceedings of the 51st {Annual} {Meeting} of the {Association} for {Computational} {Linguistics} ({Volume} 1: {Long} {Papers})}, pages 1702--1712, Sofia, Bulgaria. Association for Computational Linguistics.

\bibitem[{Frohmann et~al.(2024)Frohmann, Sterner, Vulić, Minixhofer, and Schedl}]{frohmann_segment_2024}
Markus Frohmann, Igor Sterner, Ivan Vulić, Benjamin Minixhofer, and Markus Schedl. 2024.
\newblock \href {https://doi.org/10.18653/v1/2024.emnlp-main.665} {Segment {Any} {Text}: {A} {Universal} {Approach} for {Robust}, {Efficient} and {Adaptable} {Sentence} {Segmentation}}.
\newblock In \emph{Proceedings of the 2024 {Conference} on {Empirical} {Methods} in {Natural} {Language} {Processing}}, pages 11908--11941, Miami, Florida, USA. Association for Computational Linguistics.

\bibitem[{Geng et~al.(2023{\natexlab{a}})Geng, Josifoski, Peyrard, and West}]{geng_grammar-constrained_2023}
Saibo Geng, Martin Josifoski, Maxime Peyrard, and Robert West. 2023{\natexlab{a}}.
\newblock \href {https://doi.org/10.18653/v1/2023.emnlp-main.674} {Grammar-{Constrained} {Decoding} for {Structured} {NLP} {Tasks} without {Finetuning}}.
\newblock In \emph{Proceedings of the 2023 {Conference} on {Empirical} {Methods} in {Natural} {Language} {Processing}}, pages 10932--10952, Singapore. Association for Computational Linguistics.

\bibitem[{Geng et~al.(2023{\natexlab{b}})Geng, Josifosky, Peyrard, and West}]{geng_flexible_2023}
Saibo Geng, Martin Josifosky, Maxime Peyrard, and Robert West. 2023{\natexlab{b}}.
\newblock \href {https://doi.org/10.48550/ARXIV.2305.13971} {Flexible {Grammar}-{Based} {Constrained} {Decoding} for {Language} {Models}}.
\newblock \emph{CoRR}, abs/2305.13971.
\newblock ArXiv: 2305.13971.

\bibitem[{Ghazimatin et~al.(2024)Ghazimatin, Garmash, Penha, Sheets, Achenbach, Semerci, Galvez, Tannenberg, Mantravadi, Narayanan, Kalaydzhyan, Cole, Carterette, Clifton, Bennett, Hauff, and Lalmas}]{ghazimatin_podtile_2024}
Azin Ghazimatin, Ekaterina Garmash, Gustavo Penha, Kristen Sheets, Martin Achenbach, Oguz Semerci, Remi Galvez, Marcus Tannenberg, Sahitya Mantravadi, Divya Narayanan, Ofeliya Kalaydzhyan, Douglas Cole, Ben Carterette, Ann Clifton, Paul~N. Bennett, Claudia Hauff, and Mounia Lalmas. 2024.
\newblock \href {https://doi.org/10.1145/3627673.3680081} {{PODTILE}: {Facilitating} {Podcast} {Episode} {Browsing} with {Auto}-generated {Chapters}}.
\newblock In \emph{Proceedings of the 33rd {ACM} {International} {Conference} on {Information} and {Knowledge} {Management}}, {CIKM} '24, pages 4487--4495, New York, NY, USA. Association for Computing Machinery.
\newblock Event-place: Boise, ID, USA.

\bibitem[{Ghinassi et~al.(2024)Ghinassi, Wang, Newell, and Purver}]{ghinassi_recent_2024}
Iacopo Ghinassi, Lin Wang, Chris Newell, and Matthew Purver. 2024.
\newblock \href {https://doi.org/10.18653/v1/2024.findings-emnlp.174} {Recent {Trends} in {Linear} {Text} {Segmentation}: {A} {Survey}}.
\newblock In \emph{Findings of the {Association} for {Computational} {Linguistics}: {EMNLP} 2024}, pages 3084--3095, Miami, Florida, USA. Association for Computational Linguistics.

\bibitem[{Glavaš et~al.(2021)Glavaš, Ganesh, and Somasundaran}]{glavas_training_2021}
Goran Glavaš, Ananya Ganesh, and Swapna Somasundaran. 2021.
\newblock \href {https://aclanthology.org/2021.bea-1.11} {Training and {Domain} {Adaptation} for {Supervised} {Text} {Segmentation}}.
\newblock In \emph{Proceedings of the 16th {Workshop} on {Innovative} {Use} of {NLP} for {Building} {Educational} {Applications}}, pages 110--116, Online. Association for Computational Linguistics.

\bibitem[{Grattafiori et~al.(2024)Grattafiori, Dubey, Jauhri, Pandey, Kadian, Al-Dahle, Letman, Mathur, Schelten, Vaughan, Yang, Fan, Goyal, Hartshorn, Yang, {\ldots}, and Ma}]{grattafiori_llama_2024}
Aaron Grattafiori, Abhimanyu Dubey, Abhinav Jauhri, Abhinav Pandey, Abhishek Kadian, Ahmad Al-Dahle, Aiesha Letman, Akhil Mathur, Alan Schelten, Alex Vaughan, Amy Yang, Angela Fan, Anirudh Goyal, Anthony Hartshorn, Aobo Yang, {\ldots}, and Zhiyu Ma. 2024.
\newblock \href {https://doi.org/10.48550/arXiv.2407.21783} {The {Llama} 3 {Herd} of {Models}}.
\newblock ArXiv:2407.21783 [cs].

\bibitem[{Hu et~al.(2022)Hu, Shen, Wallis, Allen-Zhu, Li, Wang, Wang, and Chen}]{hu2022lowrank}
Edward~J. Hu, Yelong Shen, Phillip Wallis, Zeyuan Allen-Zhu, Yuanzhi Li, Shean Wang, Lu~Wang, and Weizhu Chen. 2022.
\newblock \href {http://dblp.uni-trier.de/db/conf/iclr/iclr2022.html#HuSWALWWC22} {Lora: Low-rank adaptation of large language models.}
\newblock In \emph{ICLR}. OpenReview.net.

\bibitem[{Iikura et~al.(2021)Iikura, Okada, and Mori}]{iikura_improving_2021}
Riku Iikura, Makoto Okada, and Naoki Mori. 2021.
\newblock \href {https://doi.org/10.1007/978-3-030-53036-5_3} {Improving {BERT} with {Focal} {Loss} for {Paragraph} {Segmentation} of {Novels}}.
\newblock In \emph{Distributed {Computing} and {Artificial} {Intelligence}, 17th {International} {Conference}}, pages 21--30, Cham. Springer International Publishing.

\bibitem[{Jones et~al.(2003)Jones, Wolf, Gibson, Williams, Fedorenko, Reynolds, and Zissman}]{jones_measuring_2003}
Douglas~A. Jones, Florian Wolf, Edward Gibson, Elliott Williams, Evelina Fedorenko, Douglas~A. Reynolds, and Marc~A. Zissman. 2003.
\newblock \href {https://doi.org/10.21437/EUROSPEECH.2003-463} {Measuring the readability of automatic speech-to-text transcripts}.
\newblock In \emph{8th {European} {Conference} on {Speech} {Communication} and {Technology}, {EUROSPEECH} 2003 - {INTERSPEECH} 2003, {Geneva}, {Switzerland}, {September} 1-4, 2003}, pages 1585--1588. ISCA.

\bibitem[{Kamath et~al.(2025)Kamath, Ferret, Pathak, Vieillard, Merhej, Perrin, Matejovicova, Ramé, Rivière, Rouillard, Mesnard, Cideron, bastien Grill, Ramos, Yvinec, Casbon, Pot, Penchev, Liu, Visin, Kenealy, Beyer, Zhai, Tsitsulin, Busa-Fekete, Feng, Sachdeva, Coleman, Gao, Mustafa, Barr, and et~al.}]{gemma3}
Aishwarya Kamath, Johan Ferret, Shreya Pathak, Nino Vieillard, Ramona Merhej, Sarah Perrin, Tatiana Matejovicova, Alexandre Ramé, Morgane Rivière, Louis Rouillard, Thomas Mesnard, Geoffrey Cideron, Jean bastien Grill, Sabela Ramos, Edouard Yvinec, Michelle Casbon, Etienne Pot, Ivo Penchev, Gaël Liu, Francesco Visin, Kathleen Kenealy, Lucas Beyer, Xiaohai Zhai, Anton Tsitsulin, Robert Busa-Fekete, Alex Feng, Noveen Sachdeva, Benjamin Coleman, Yi~Gao, Basil Mustafa, Iain Barr, and Emilio~Parisotto et~al. 2025.
\newblock \href {http://arxiv.org/abs/2503.19786} {Gemma 3 technical report}.

\bibitem[{Koshorek et~al.(2018)Koshorek, Cohen, Mor, Rotman, and Berant}]{koshorek_text_2018}
Omri Koshorek, Adir Cohen, Noam Mor, Michael Rotman, and Jonathan Berant. 2018.
\newblock \href {https://doi.org/10.18653/v1/N18-2075} {Text {Segmentation} as a {Supervised} {Learning} {Task}}.
\newblock In \emph{Proceedings of the 2018 {Conference} of the {North} {American} {Chapter} of the {Association} for {Computational} {Linguistics}: {Human} {Language} {Technologies}, {Volume} 2 ({Short} {Papers})}, pages 469--473, New Orleans, Louisiana. Association for Computational Linguistics.

\bibitem[{Lai et~al.(2016)Lai, Farrús, and Moore}]{lai_automatic_2016}
Catherine Lai, Mireia Farrús, and Johanna~D. Moore. 2016.
\newblock \href {https://doi.org/10.21437/Interspeech.2016-992} {Automatic {Paragraph} {Segmentation} with {Lexical} and {Prosodic} {Features}}.
\newblock In \emph{Interspeech 2016}, pages 1034--1038.
\newblock ISSN: 2958-1796.

\bibitem[{Lai et~al.(2020)Lai, Farrús, and Moore}]{lai_integrating_2020}
Catherine Lai, Mireia Farrús, and Johanna~D. Moore. 2020.
\newblock \href {https://doi.org/10.1016/j.specom.2020.04.007} {Integrating lexical and prosodic features for automatic paragraph segmentation}.
\newblock \emph{Speech Communication}, 121:44--57.

\bibitem[{Lukasik et~al.(2020)Lukasik, Dadachev, Papineni, and Simões}]{lukasik_text_2020}
Michal Lukasik, Boris Dadachev, Kishore Papineni, and Gonçalo Simões. 2020.
\newblock \href {https://doi.org/10.18653/v1/2020.emnlp-main.380} {Text {Segmentation} by {Cross} {Segment} {Attention}}.
\newblock In \emph{Proceedings of the 2020 {Conference} on {Empirical} {Methods} in {Natural} {Language} {Processing} ({EMNLP})}, pages 4707--4716, Online. Association for Computational Linguistics.

\bibitem[{Lundberg and Ribeiro(2023)}]{lundberg_ribeiro_2023_token_healing}
Scott Lundberg and Marco~Tulio Ribeiro. 2023.
\newblock The art of prompt design: Prompt boundaries and token healing.
\newblock \emph{Towards Data Science (Medium)}.

\bibitem[{Minixhofer et~al.(2023)Minixhofer, Pfeiffer, and Vulić}]{minixhofer_wheres_2023}
Benjamin Minixhofer, Jonas Pfeiffer, and Ivan Vulić. 2023.
\newblock \href {https://doi.org/10.18653/v1/2023.acl-long.398} {Where's the {Point}? {Self}-{Supervised} {Multilingual} {Punctuation}-{Agnostic} {Sentence} {Segmentation}}.
\newblock In \emph{Proceedings of the 61st {Annual} {Meeting} of the {Association} for {Computational} {Linguistics} ({Volume} 1: {Long} {Papers})}, pages 7215--7235, Toronto, Canada. Association for Computational Linguistics.

\bibitem[{Pappu and Stent(2015)}]{pappu_automatic_2015}
Aasish Pappu and Amanda Stent. 2015.
\newblock \href {https://doi.org/10.21437/Interspeech.2015-542} {Automatic formatted transcripts for videos}.
\newblock In \emph{Interspeech 2015}, pages 2514--2518.
\newblock ISSN: 2958-1796.

\bibitem[{Retkowski and Waibel(2024)}]{retkowski_text_2024}
Fabian Retkowski and Alexander Waibel. 2024.
\newblock \href {https://aclanthology.org/2024.eacl-long.25} {From {Text} {Segmentation} to {Smart} {Chaptering}: {A} {Novel} {Benchmark} for {Structuring} {Video} {Transcriptions}}.
\newblock In \emph{Proceedings of the 18th {Conference} of the {European} {Chapter} of the {Association} for {Computational} {Linguistics}, {EACL} 2024 - {Volume} 1: {Long} {Papers}, {St}. {Julian}'s, {Malta}, {March} 17-22, 2024}, pages 406--419. Association for Computational Linguistics.

\bibitem[{Salimbajevs and Ikauniece(2017)}]{salimbajevs_system_2017}
Askars Salimbajevs and Indra Ikauniece. 2017.
\newblock System for {Speech} {Transcription} and {Post}-{Editing} in {Microsoft} {Word}.
\newblock In \emph{Interspeech 2017}, pages 825--826.
\newblock ISSN: 2958-1796.

\bibitem[{Shugrina(2010)}]{shugrina_formatting_2010}
Maria Shugrina. 2010.
\newblock \href {https://aclanthology.org/N10-1023/} {Formatting {Time}-{Aligned} {ASR} {Transcripts} for {Readability}}.
\newblock In \emph{Human {Language} {Technologies}: {The} 2010 {Annual} {Conference} of the {North} {American} {Chapter} of the {Association} for {Computational} {Linguistics}}, pages 198--206, Los Angeles, California. Association for Computational Linguistics.

\bibitem[{Sporleder and Lapata(2004)}]{sporleder_automatic_2004}
Caroline Sporleder and Mirella Lapata. 2004.
\newblock \href {https://aclanthology.org/W04-3210/} {Automatic {Paragraph} {Identification}: {A} {Study} across {Languages} and {Domains}}.
\newblock In \emph{Proceedings of the 2004 {Conference} on {Empirical} {Methods} in {Natural} {Language} {Processing} , {EMNLP} 2004, {A} meeting of {SIGDAT}, a {Special} {Interest} {Group} of the {ACL}, held in conjunction with {ACL} 2004, 25-26 {July} 2004, {Barcelona}, {Spain}}, pages 72--79. ACL.

\bibitem[{Stark(1988)}]{stark_what_1988}
Heather~A. Stark. 1988.
\newblock \href {https://doi.org/10.1080/01638538809544704} {What do paragraph markings do?}
\newblock \emph{Discourse Processes}, 11(3):275--303.
\newblock Publisher: Routledge \_eprint: https://doi.org/10.1080/01638538809544704.

\bibitem[{Tündik et~al.(2018)Tündik, Szaszák, Gosztolya, and Beke}]{tundik_user-centric_2018}
Máté~Ákos Tündik, György Szaszák, Gábor Gosztolya, and András Beke. 2018.
\newblock \href {https://doi.org/10.21437/Interspeech.2018-1352} {User-centric {Evaluation} of {Automatic} {Punctuation} in {ASR} {Closed} {Captioning}}.
\newblock In \emph{Interspeech 2018}, pages 2628--2632.
\newblock ISSN: 2958-1796.

\bibitem[{Wicks and Post(2021)}]{wicks_unified_2021}
Rachel Wicks and Matt Post. 2021.
\newblock \href {https://doi.org/10.18653/v1/2021.acl-long.309} {A unified approach to sentence segmentation of punctuated text in many languages}.
\newblock In \emph{Proceedings of the 59th {Annual} {Meeting} of the {Association} for {Computational} {Linguistics} and the 11th {International} {Joint} {Conference} on {Natural} {Language} {Processing} ({Volume} 1: {Long} {Papers})}, pages 3995--4007, Online. Association for Computational Linguistics.

\bibitem[{Yang et~al.(2025)Yang, Yang, Zhang, Hui, Zheng, Yu, Li, Liu, Huang, Wei, Lin, Yang, Tu, Zhang, Yang, Yang, Zhou, Lin, Dang, Lu, Bao, Yang, Yu, Li, Xue, Zhang, Zhu, Men, Lin, Li, Tang, Xia, Ren, Ren, and et~al.}]{qwen25}
An~Yang, Baosong Yang, Beichen Zhang, Binyuan Hui, Bo~Zheng, Bowen Yu, Chengyuan Li, Dayiheng Liu, Fei Huang, Haoran Wei, Huan Lin, Jian Yang, Jianhong Tu, Jianwei Zhang, Jianxin Yang, Jiaxi Yang, Jingren Zhou, Junyang Lin, Kai Dang, Keming Lu, Keqin Bao, Kexin Yang, Le~Yu, Mei Li, Mingfeng Xue, Pei Zhang, Qin Zhu, Rui Men, Runji Lin, Tianhao Li, Tianyi Tang, Tingyu Xia, Xingzhang Ren, Xuancheng Ren, and et~al. 2025.
\newblock \href {http://arxiv.org/abs/2412.15115} {Qwen2.5 technical report}.

\bibitem[{Yoo and Kim(2024)}]{yoo_improving_2024}
Byunghwa Yoo and Kyung-Joong Kim. 2024.
\newblock \href {https://doi.org/10.1016/j.nlp.2024.100061} {Improving paragraph segmentation using {BERT} with additional information from probability density function modeling of segmentation distances}.
\newblock \emph{Natural Language Processing Journal}, 6:100061.

\bibitem[{Zadrozny and Jensen(1991)}]{zadrozny_semantics_1991}
Wlodek Zadrozny and Karen Jensen. 1991.
\newblock Semantics of paragraphs.
\newblock \emph{Comput. Linguist.}, 17(2):171--209.
\newblock Place: Cambridge, MA, USA Publisher: MIT Press.

\bibitem[{Zhuo et~al.(2023)Zhuo, Murata, and Ma}]{zhuo_auxiliary_2023}
Binggang Zhuo, Masaki Murata, and Qing Ma. 2023.
\newblock \href {https://doi.org/10.1587/transinf.2022EDP7083} {Auxiliary {Loss} for {BERT}-{Based} {Paragraph} {Segmentation}}.
\newblock \emph{IEICE Transactions on Information and Systems}, E106.D(1):58--67.

\end{thebibliography}

\section{Language Resource References}
\label{lr:ref}
\bibliographystylelanguageresource{lrec2026-natbib}
\bibliographylanguageresource{languageresource}

\clearpage

\appendix

\section{Prompt Template}

\begin{figure*}[h]
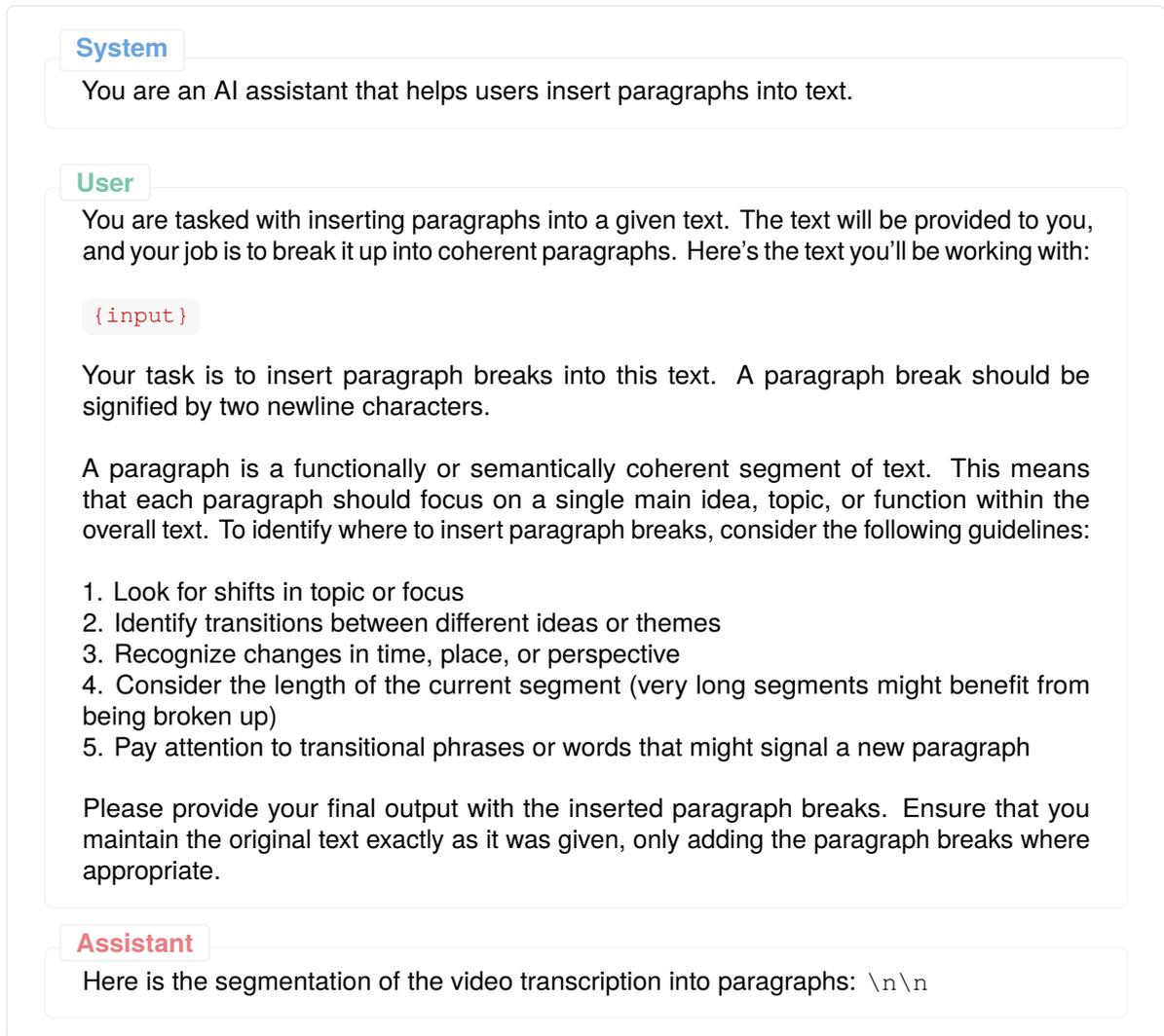

    \centering
    \begin{minipage}[h]{0.98\textwidth}
        \vspace{0pt}
        \begin{tcolorbox}[custombox, width=\linewidth]
            \begin{tcolorbox}[custombox, title={\textcolor{systemcolor}{System}}, colframe=gray!10]
                You are an AI assistant that helps users insert paragraphs into text.
            \end{tcolorbox}
            
            \vspace{0.5mm}
            
            \begin{tcolorbox}[custombox, title={\textcolor{usercolor}{User}}, colframe=gray!10]
                You are tasked with inserting paragraphs into a given text. The text will be provided to you, and your job is to break it up into coherent paragraphs. Here's the text you'll be working with:\\

                \formatvariable{\{input\}}\\

                Your task is to insert paragraph breaks into this text. A paragraph break should be signified by two newline characters.\\

                A paragraph is a functionally or semantically coherent segment of text. This means that each paragraph should focus on a single main idea, topic, or function within the overall text. To identify where to insert paragraph breaks, consider the following guidelines:\\

                1. Look for shifts in topic or focus

                2. Identify transitions between different ideas or themes

                3. Recognize changes in time, place, or perspective

                4. Consider the length of the current segment (very long segments might benefit from being broken up)

                5. Pay attention to transitional phrases or words that might signal a new paragraph\\

                Please provide your final output with the inserted paragraph breaks. Ensure that you maintain the original text exactly as it was given, only adding the paragraph breaks where appropriate.
            \end{tcolorbox}
            \begin{tcolorbox}[custombox, title={\textcolor{assistantcolor}{Assistant}}, colframe=gray!10]
                Here is the segmentation of the video transcription into paragraphs:
                \texttt{\textbackslash n\textbackslash n}
            \end{tcolorbox}

        \end{tcolorbox}
        \caption{Prompt Template for Paragraph Insertion with LLMs}
        \label{fig:paragraph-segmentation-prompt}
    \end{minipage}
\end{figure*}

\Cref{fig:paragraph-segmentation-prompt} shows the prompt template used in both the naive and constrained decoding settings. We also employ prompt prefilling (i.e., including a stub assistant reply) to better guide the model toward generating paragraph-structured output.

\section{LLM Hyperparameters}

\paragraph{Greedy Decoding.} For all LLM-based inference, we apply \textit{greedy decoding}. This applies to both the naïve baseline as well as the constrained decoding setup. In the constrained decoding case, however, the output space at sentence boundaries is explicitly restricted to punctuation tokens only.

\paragraph{Context Handling.} All LLaMA-based models used in our experiments support a context window of up to 128K tokens. This enables processing of long-form spoken content, including multi-hour YouTube transcripts from \textsc{YTSegPara}. While some transcripts exceed the model’s context limit, we employ chapter-wise processing, allowing all documents to be processed.

\paragraph{LoRA Fine-Tuning.} We fine-tune LLaMA~3.1-8B-Instruct \cite{grattafiori_llama_2024} using LoRA \cite{hu2022lowrank} on the \textsc{TEDPara} training set with the prompt template from \Cref{fig:paragraph-segmentation-prompt}. To address the class imbalance between continuation and break tokens, we experiment with weighted cross-entropy, upweighting \texttt{\textbackslash n\textbackslash n} break tokens by a factor $w \in \{1.0, 1.5, 2.0\}$ (see \Cref{tab:token_weight}). The best checkpoint is selected by validation loss. All hyperparameters are listed in \Cref{tab:lora_hyperparameters}.

\section{MiniSeg Hyperparameters}

For training MiniSeg on the \textsc{TEDPara} and \textsc{YTSegPara} datasets, we largely follow the hyperparameter configuration established in the original work and implementation by \citet{retkowski_text_2024}. The full set of hyperparameters used is summarized in \Cref{tab:hyperparameters}. A key aspect of the training setup involves the weighting scheme for the weighted cross-entropy loss. For \textsc{TEDPara}, we retain the original weighting of $[1.2]$ as proposed. In contrast, \textsc{YTSegPara} involves a hierarchical segmentation task with three distinct classes, allowing for class-specific weighting. Based on validation experiments, a class weight configuration of $[1, 1.5, 2]$ was found to be effective.

\section{Segmentation Evaluation}

We utilize the \texttt{segeval}\footnote{\url{https://segeval.readthedocs.io/}} library \cite{fournier_evaluating_2013} to calculate segmentation evaluation metrics, such as $P_k$ and Boundary Similarity, using the default parameter configurations in both cases.

\section{Human Evaluation}
We conducted a two-part human evaluation on the TEDPara test dataset with 8 participants (4 per subtask; 30 trials each), comparing random and rule-based baselines, LLM-generated segmentations (with and without PBR), and human-annotated references. In Subtask 1, annotators performed randomized, blind pairwise A/B comparisons of segmentations with three response options: A, B, or tie; to mitigate position bias, presentation order was randomized, and model–text pairs were sampled online using inverse-frequency weighting to avoid per-participant repeats and ensure balanced coverage. In Subtask 2, participants rated individual segmentations on a 5-point Likert scale, with each trial assigned by a balanced sampler that inversely weighted overrepresented models and prioritized unseen texts, again ensuring broad and nonredundant evaluation coverage.

\clearpage

  \begin{table*}[htb]
  \centering
  \begin{tabular}{lr}
  \toprule
  \textbf{Hyperparameter} & \textbf{Value} \\
  \midrule
  Base Model         & LLaMA 3.1-8B-Instruct \\
  LoRA Rank ($r$)    & 16 \\
  LoRA Alpha ($\alpha$) & 32 \\
  LoRA Dropout       & 0.05 \\
  LoRA Target        & All linear layers \\
  Max Sequence Length & 8{,}192 \\
  Epochs             & 3 \\
  Effective Batch Size & 32 \\
  Learning Rate      & $1 \times 10^{-4}$ \\
  LR Schedule        & Cosine \\
  Warmup Ratio       & 0.1 \\
  Precision          & bfloat16 \\
  Checkpoint Selection & Best validation loss \\
  Break Token Weight ($w$) & $\{1.0, 1.5, 2.0\}$ \\
  \bottomrule
  \end{tabular}
  \caption{LoRA fine-tuning hyperparameters.}
  \label{tab:lora_hyperparameters}
  \end{table*}

\begin{table*}[htb]
    \centering
    \begin{tabular}{p{4.5cm}rr}
        \toprule
        \textbf{Hyperparameter} & \textsc{TEDPara} & \textsc{YTSegPara} \\
        \midrule
        Sentence Encoder & \multicolumn{2}{r}{\texttt{\small all-MiniLM-L12-v2}} \\
        Loss Function & \multicolumn{2}{r}{Weighted Binary Cross-Entropy} \\
        Learning Rate & \multicolumn{2}{r}{$2.5 \times 10^{-5}$} \\
        Batch Size & \multicolumn{2}{r}{115,000 Tokens} \\
        Epochs & \multicolumn{2}{r}{15} \\
        Learning Rate Schedule & \multicolumn{2}{r}{Cosine} \\
        Optimizer & \multicolumn{2}{r}{AdamW} \\
        Dropout Rate & \multicolumn{2}{r}{0.1} \\
        Gradient Sampling Rate & \multicolumn{2}{r}{0.5} \\
        Cross-Entropy Class Weights & $[1.2]$ & $[1, 1.5, 2]$ \\
        \bottomrule
    \end{tabular}
    \caption{MiniSeg Hyperparameters for Training on \textsc{TEDPara} and \textsc{YTSegPara}}
    \label{tab:hyperparameters}
\end{table*}



\end{document}